\documentclass{article}

\usepackage[preprint]{neurips_2025}

\usepackage[utf8]{inputenc} 
\usepackage[T1]{fontenc}    
\usepackage{hyperref}       
\usepackage{url}            
\usepackage{booktabs}       
\usepackage{amsfonts}       
\usepackage{nicefrac}       
\usepackage{microtype}      
\usepackage{xcolor}         
\usepackage{amsmath}
\usepackage{graphicx}
\usepackage{subcaption}

\usepackage{multirow}  
\usepackage{array}     
\usepackage{siunitx}   

\usepackage{algorithm}  
\usepackage{algpseudocode}  
\usepackage{tcolorbox}  

\newcommand{\pycomment}[1]{\textcolor{gray}{\texttt{\# #1}}}

\title{Cognitive Duality for Adaptive Web Agents}

\author{
  Jiarun Liu, Chunhong Zhang, Zheng Hu \\
  Beijing University of Posts and Telecommunications \\
  \texttt{\{liujiarun01, zhangch, huzheng\}@bupt.edu.cn}
}

\begin{document}

\maketitle

\begin{abstract}
    Web navigation represents a critical and challenging domain for evaluating artificial general intelligence (AGI), demanding complex decision-making within high-entropy, dynamic environments with combinatorially explosive action spaces. Current approaches to building autonomous web agents either focus on offline imitation learning or online exploration, but rarely integrate both paradigms effectively. Inspired by the dual-process theory of human cognition, we derive a principled decomposition into fast (\texttt{System 1}) and slow (\texttt{System 2}) cognitive processes. This decomposition provides a unifying perspective on existing web agent methodologies, bridging the gap between offline learning of intuitive reactive behaviors and online acquisition of deliberative planning capabilities. We implement this framework in \texttt{CogniWeb}, a modular agent architecture that adaptively toggles between fast intuitive processing and deliberate reasoning based on task complexity. Our evaluation on WebArena demonstrates that \texttt{CogniWeb} achieves competitive performance (43.96\% success rate) while maintaining significantly higher efficiency (75\% reduction in token usage).
\end{abstract}
\begin{figure}[htbp]
    \centering
    \includegraphics[width=0.72\textwidth]{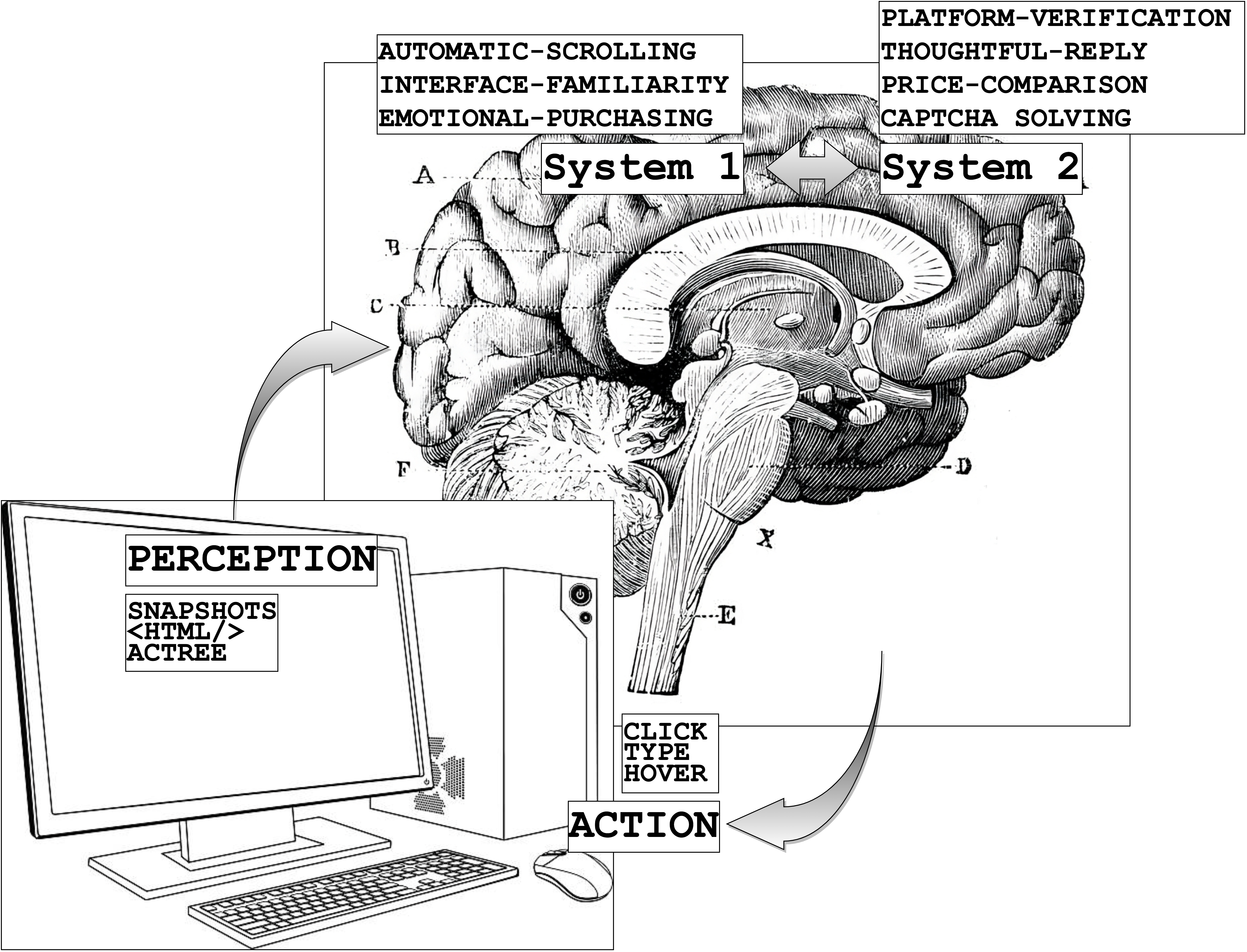}
    \caption{In computer using task such as web tasks, people tend to employ two distinct modes for processing information, aligning with dual-system theory. The fast thinking mode relies on intuition and heuristics, responsible for actions such as clicking buttons based on interface similarity; while the slow thinking mode employs logic, activated during text input or conditional filtered searches.}
    \label{fig:diagram_1}
\end{figure}

\section{Introduction}

Computer Using Agents (CUAs) \citep{cua2025} are experiencing unprecedented growth in research interest and real-world applications, as they promise to automate increasingly complex tasks across digital interfaces and enhance human productivity. Web navigation represents one of the most practical yet challenging domains for advancing artificial general intelligence (AGI). The complexity of web environments—characterized by high entropy, dynamic content, and vast action spaces—provides a rigorous testbed for agent capabilities. Recent advances in large language models (LLMs) have accelerated progress in web agents, with numerous approaches emerging to address the challenges of web navigation from different angles \citep{zhou2023webarena,deng2023mind2web,gur2023real,zhang2025webpilot,lu2024weblinx}. However, these approaches typically emphasize either offline imitation learning from human demonstrations or online learning through environmental interaction, with limited integration between these complementary paradigms.

In this paper, we address this limitation by introducing a theoretical framework that unifies offline and online learning approaches through the lens of dual-process cognitive theory \citep{kahneman2011thinking}. While \citet{lin2023swiftsage} previously implemented a dual-process approach for general interactive reasoning tasks by integrating behavior cloning with large language models, our work is the first to systematically formalize and apply this cognitive framework specifically to web navigation, deriving principled mechanisms for switching between fast and slow thinking modes in this domain. Our key insight is that optimal web task inherently requires balancing two distinct cognitive processes: a fast, intuitive system for familiar patterns and a slow, deliberate system for complex planning—analogous to Kahneman's \textit{"thinking fast and slow."} By formalizing these processes mathematically, we derive a principled decomposition of the web navigation problem that clarifies when and how to leverage each cognitive mode. Our primary contributions are:
\begin{itemize}
    \item \textbf{Dual-System Formulation}: We establish a mathematical framework that formalizes web navigation as a complexity-weighted optimization problem, demonstrating how web environments provide an ideal testbed for general intelligence. We derive a principled decomposition of web navigation into fast intuitive processing and deliberative reasoning, providing the first formal integration of dual-process cognitive theory into computer using agent design.
    \item \textbf{Unifying Framework}: Our approach bridges the gap between offline imitation learning and online exploration, establishing connections between seemingly disparate methodologies in current web agent research and providing a unified perspective.
    \item \textbf{Implementation and Validation}: We implement our theoretical framework, a modular agent architecture that adaptively toggles between two systems based on task complexity, demonstrating competitive performance on WebArena while maintaining substantially higher efficiency than pure reasoning approaches.
\end{itemize}

Our work contributes to the broader goal of developing more capable and efficient web agents by establishing a theoretically grounded framework that unifies existing approaches and clarifies pathways for future research. By formalizing the complementary nature of fast intuitive processing and slow deliberative reasoning, we enable more effective integration of offline and online learning methods, paving the way for web agents that can scale to increasingly complex real-world tasks.

\section{Theory}
\label{sec:theory}
\subsection{Web Navigation as a Complex Testbed for General Intelligence}

\citet{legg2007universal} proposed an influential formalization, describing intelligence as \textit{a measure of an agent's ability to achieve goals across various environments.} This definition inspired theoretical frameworks like AIXI, which combines Solomonoff induction \cite{legg1997solomonoff} and reinforcement learning to create a theoretical model of artificial general intelligence (AGI). According to the Legg-Hutter framework, general intelligence can be formally represented as
\begin{equation}
\Upsilon(\pi) = \sum_{\mu} 2^{-K(\mu)} V^{\pi}_{\mu} 
\end{equation}
where $\Upsilon(\pi)$ represents the general intelligence measure of agent $\pi$, $K(\mu)$ is the Kolmogorov complexity of environment $\mu$ (a complexity-based weighting factor), $V^{\pi}_{\mu}$ is the expected cumulative reward of agent $\pi$ in environment $\mu$. 

The web environment can be expressed as a graph $G = (V, E)$, where $V$ represents all possible webpages and $E$ represents links between pages. The Kolmogorov complexity $K(\mu)$ of the web is extremely high, approximately $K(\mu) \approx \log_2|V| + |V|\cdot H(E|V)$ where $H(E|V)$ represents the conditional entropy of the edge set given the vertex set \citep{li2008introduction}. Considering that the environment contains billions of nodes and trillions of edges, its \textit{complexity} far exceeds that of most traditional testbeds. The entropy of the web environment can be quantified with information theory as
\begin{equation}
H(web) = -\sum_{i=1}^{|V|} P(v_i) \log_2 P(v_i) + \sum_{i=1}^{|V|} P(v_i) \cdot H(E|v_i)
\end{equation}
where $|V|$ is the total number of webpages, $P(v_i)$ is the probability of visiting a specific webpage, and $H(E|v_i)$ is the entropy of the link structure given webpage $v_i$. Traditional reinforcement learning environments and web environments differ significantly in their entropy magnitudes. Taking Atari games as an example, its state space is approximately $10^{10}$ \citep{mnih2015human, bellemare2013arcade}. In contrast, the web environment's state space exceeds $10^{20}$ \citep{broder2000graph}. Hence we have
\begin{equation}
\frac{H(web)}{H(Atari)} \approx \frac{|V|{web} \cdot \log_2(|E|{web}/|V|{web})}{|V|{Atari} \cdot \log_2(|E|{Atari}/|V|{Atari})} \approx \frac{10^9 \cdot \log_2(10^{12}/10^9)}{10^4 \cdot \log_2(10^5/10^4)} > 10^5
\end{equation}

Additionally, the web possesses \textit{dynamic} characteristics. Websites structures and their inner contents continuously update, which can be mathematically modeled as $P(S_{t+1}|S_t) \neq \delta(S_{t+1} - S_t)$, where the Dirac indicator $\delta$ tells the environment is non-static \citep{puterman2014markov}. That is to say, maximizing the general intelligence measure $\Upsilon_T(\pi)$ in a time-varying web environment is equivalent to maximizing the weighted average of discounted cumulative rewards (value) across all possible states and time points, where weights are determined by the dynamic complexity of the environment.  This formal perspective emphasizes the unique value of \textit{web navigation as a test of AGI}. It requires agents to conduct effective exploration and planning in environments characterized by extremely high entropy, dynamic changes, and combinatorially explosive action spaces.

\subsection{Deriving Fast and Slow Thinking Mechanisms for Web Navigation}
\label{sec2.2}

Building upon our analysis of web navigation complexity, we now derive a formal optimization objective for web navigation agents. Starting from the time-varying general intelligence measure, we can express the learning objective $\mathcal{J}(\theta)$ for a parameterized policy $\pi_\theta$ as
\begin{equation}
\Upsilon_T(\pi_\theta) = \mathbb{E}{\mu,t,\tau}\left[\gamma(t) \cdot 2^{-K(\mu_t)} \cdot R(\tau)\right] \equiv \mathcal{J}(\theta)
\end{equation}
For LLM-based agents, actions are generated through token probabilities \citep{sumers2023cognitive}, connecting our reward function to the language model's output distribution
\begin{equation}
\pi_{\theta}=p_\theta(\tau|\mu) = \prod_{t=0}^{T} p_\theta(a_t|s_{1:t}, a_{1:t-1}, g)
\end{equation}
As emphasized in \cite{yao2022webshop} and other research, we reiterate that $s$ represents the webpage state, $a$ is an action set from various \textit{modalities} (such as language instructions and APIs \citep{lu2024weblinx, gur2023real}, or visual pixel coordinates \citep{lee2023pix2struct}), and different \textit{granularities} (theoretically, any website can map to a set of $M \gg 2$ APIs to satisfy all its services, or alternatively, one can interact with a web element through just two numerical coordinates, e.g., $x, y$ values for pixel-based actions). We use $g$ to represent the high-level abstract intent (a.k.a. task or instruction).

As demonstrated by test-time scaling law \citep{snell2024scalingllmtesttimecompute}, there exists a systematic relationship between model parameter size, inference-time computation, and performance on complex reasoning tasks. For $\pi_\theta$ the general intelligence measure $\Upsilon_T(\pi_\theta)$ monotonically increases with both $|\theta|$ and the inference-time computation budget (measured by reasoning length $r$) until saturation. Consider a time-varying environment $\mu_t$ and the associated optimal transition $\hat{p}_\theta(a_t|s_{1:t}, a_{1:t-1}, g)$, the test-time scaling law provides an empirical guarantee $ \exists\, \tilde{\theta} \in \Theta,\, \tilde{r} \in R$, such that
\begin{equation}
\hat{p}_\theta(a_t|s_{1:t}, a_{1:t-1}, g) = \prod_{i=1}^{\tilde{r}} p_{\tilde{\theta}}(\text{token}_i|\text{token}_{1:i-1},s_{1:t}, a_{1:t-1}, g) \quad 
\label{eq_trans}
\end{equation}
Therefore, we can reformulate our policy optimization objective as finding the optimal combination of parameter settings and reasoning lengths, and rearrange the configurations by sorting $r_i$ and $\theta_i$
\begin{align}
\hat{\pi}_{\theta} = \hat{p}_\theta(\tau|\mu) = \prod_{j=1}^{N} (\prod_{i=1}^{r_j} p_{\theta_j}(\cdot)) = \prod^{r_1} p_{\theta_1}(\cdot) \times \prod^{r_2} p_{\theta_2}(\cdot) \times \ldots
\label{eq_10}
\end{align}
where $p_{\theta_i}(\cdot)$ is derived from \eqref{eq_trans}, and $(\theta_j,r_j)$ is sorted in ascending order. Turning our focus to web navigation, we can make this formulation more practical by analyzing the distribution of intents difficulty in real-world web environments.

\textbf{\textit{Observation}}. \textit{The difficulty of a web navigation task $\Upsilon_T(\pi_\theta)$ is approximately proportional to the number of steps required to complete it.} 

\begin{figure}[htbp]
  \centering
  \begin{minipage}{0.84\textwidth}
    \begin{subfigure}[b]{0.48\textwidth}
      \includegraphics[width=\linewidth]{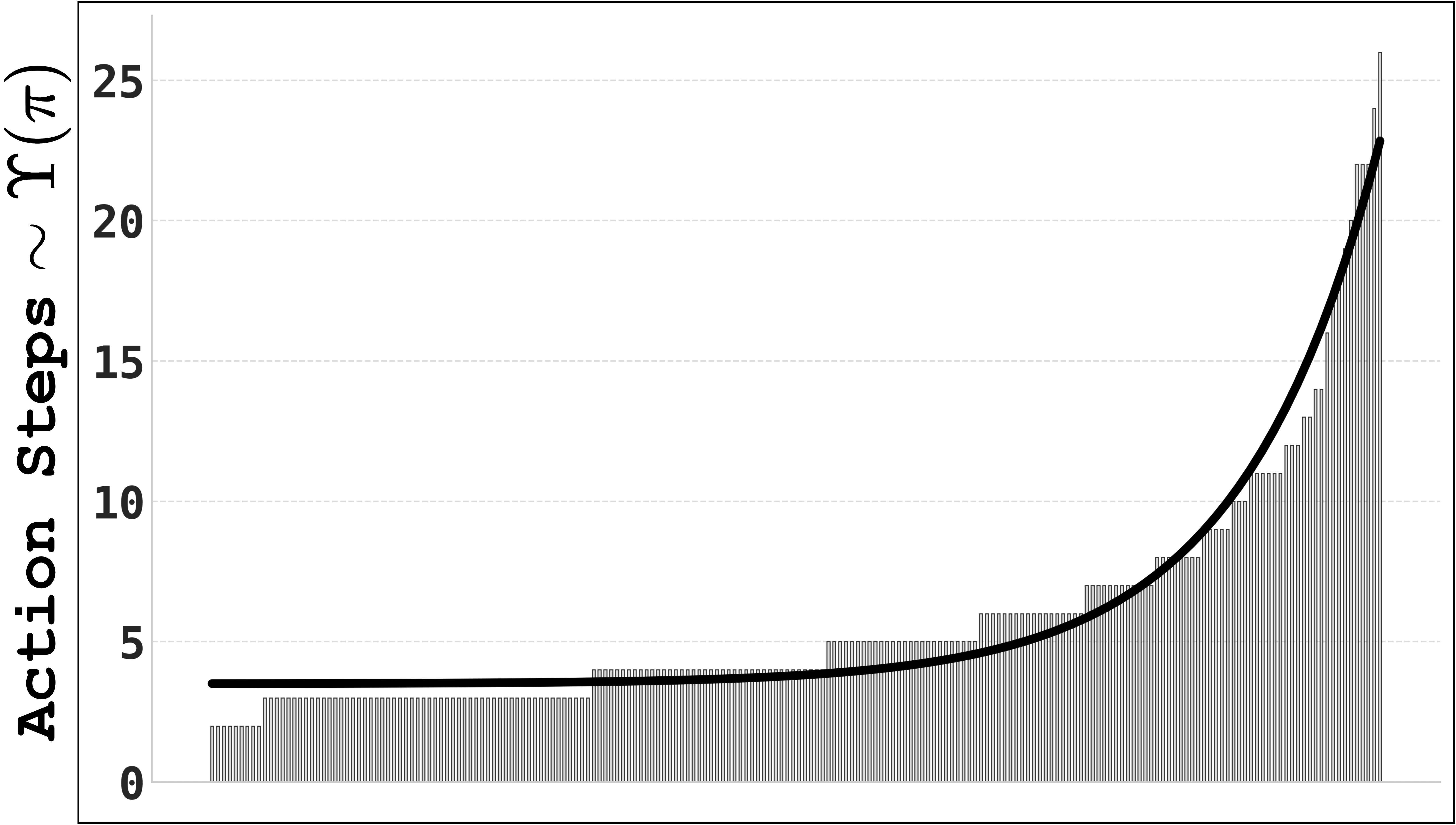}
      \caption{Sampled task steps distribution from Mind2Web and WebArena human trajectories. We can derive our \textit{\textbf{observation}} from the shape of the fitted curve, which provides a rationale for simplifying and merging parameters in the subsequent steps.}
      \label{fig:sub1}
    \end{subfigure}
    \hfill
    \begin{subfigure}[b]{0.46\textwidth}
      \includegraphics[width=\linewidth]{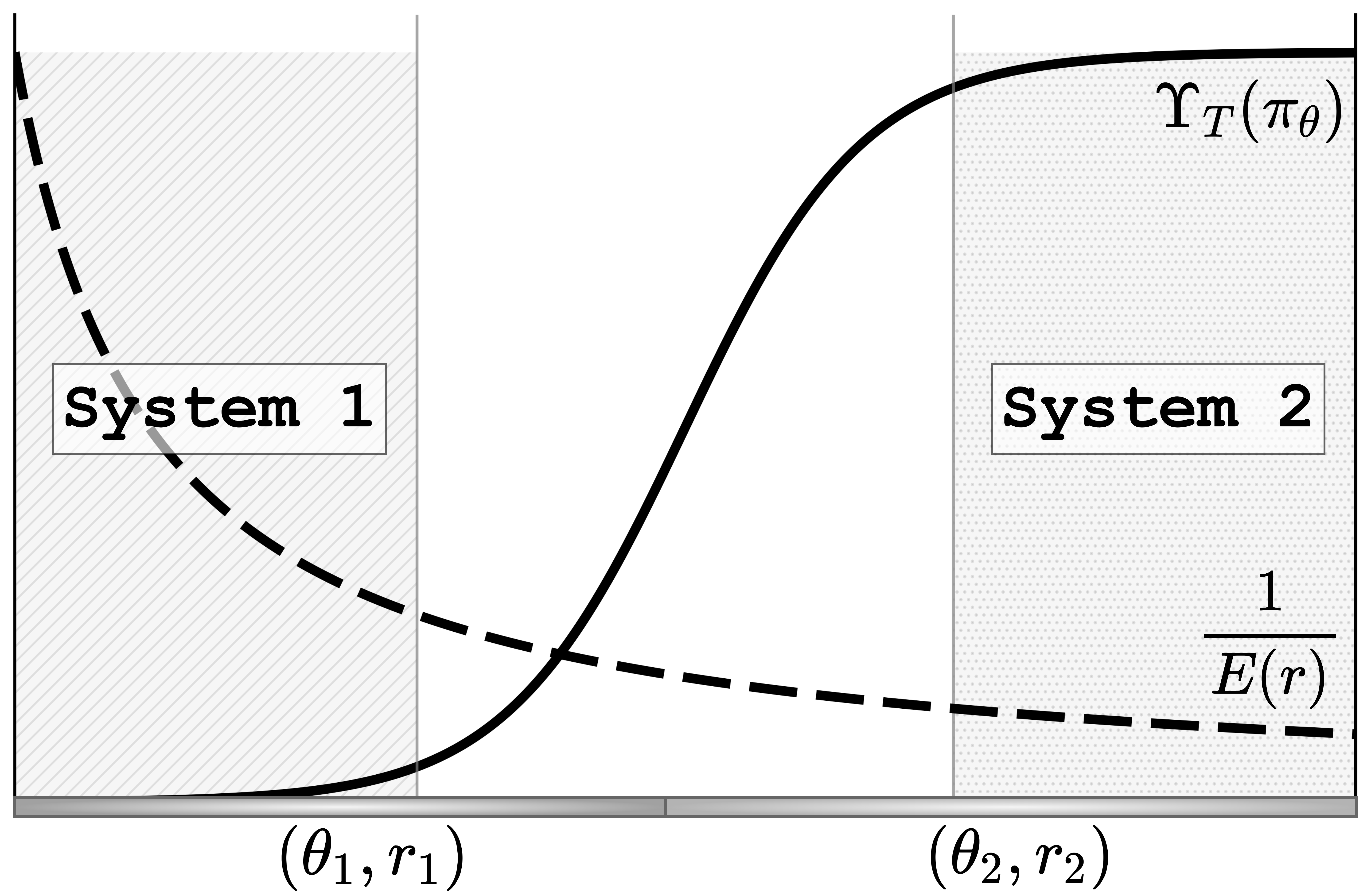}
      \caption{Capability-Efficiency Curve. In the \texttt{system-1} region, efficiency is improved when task difficulty is low, whereas in the \texttt{system-2} region, higher intelligence is achieved by sacrificing efficiency.}
      \label{fig:sub2}
    \end{subfigure}
  \end{minipage}
  \caption{Task complexity analysis and capability-efficiency trade-offs.}
  \label{fig:twopics}
\end{figure}

We then analyzed task step distributions from \citet{deng2023mind2web} and \citet{zhou2023webarena} human collected trajectories based on this observation. As illustrated in Figure \ref{fig:sub1}, the distribution of action steps follows an approximately exponential arise, indicating that while most intents are relatively simple, there exists a long tail of complex tasks requiring numerous steps. It suggests that \textit{real-world high-level intents naturally cluster into different complexity tiers, particularly distinguishing between simple and complex ones.} Correspondingly, as shown in Figure \ref{fig:sub2}, we can analyze how policy capability $\Upsilon_T(\pi_\theta)$ and computational efficiency $E(r)^{-1}$ vary with $(\theta,r)$. The capability curve follows a sigmoid-like distribution, while the efficiency decreases with increased $(\theta,r)$.

Above analysis reveals that rather than selecting a single intermediate value $(\theta, r)$ as a fixed policy configuration, we can achieve better performance by identifying two distinct configurations $(\theta_1, r_1)$ and $(\theta_2, r_2)$ that \textit{better anchor the lower and upper boundaries of the capability-efficiency curve}, thus more effectively addressing both simple and complex tasks.

We therefore restructure our policy parameters as $\theta = \theta_1 \cup \theta_2$ with abuse, where $\theta_1$ corresponds to a configuration optimized for efficiency with minimal reasoning length $r_1$ (ideally $r_1=\text{len}(a_t)$ for direct output without reasoning steps), while $\theta_2$ leverage arbitrary reasoning length $r_2$ to handle complex tasks. As illustrated in Figure \ref{fig:sub2}, the regions on either side of the threshold correspond to the respective \textit{"domains of competence"} for $p_{\theta_1}$ and $p_{\theta_2}$. Based on \eqref{eq_10}, we can identify a median $(\theta', r')$ that separates these two regimes, yielding our estimated transition distribution
\begin{equation}
\hat{\pi}_{\theta} = \prod_{j=1}^{N_1} (\prod_{i=1}^{r_j \le r'} p_{\theta_j \le \theta'}(\cdot)) \times  \prod_{j=1}^{N_2} (\prod_{i=1}^{r_j > r'} p_{\theta_j > \theta'}(\cdot))
\overset{d}{\approx} \underbrace{\prod^{N_1}(\prod^{r_1} p_{\theta_1}(\cdot))}_{\texttt{System 1's outputs }} \times \underbrace{\prod^{N_2}(\prod^{r_2} p_{\theta_2}(\cdot))}_{\texttt{System 2's outputs }}
\end{equation}
where $N_1+N_2=N$. This decomposition reveals that optimal decision-making in web environments involves balancing two distinct cognitive processes, one that is immediate and based on minimal information, and the other that considers the long-term consequences of decisions, drawing on past experiences and reasoning over multiple steps. We can therefore represent our policy as a mixture of two sub-policies 
\begin{equation}
p_\theta(a_t|s_{1:t},a_{1:t-1},g) = \lambda_t \cdot \pi_1(a_t|s_t, g) + (1-\lambda_t) \cdot \pi_2(a_t|s_t, h_t, g)
\end{equation}
with switch $\lambda_t \in [0,1]$ determining which system dominates at each timestep, and $h_t$ demonstrates working memory \citep{BADDELEY197447} or episodic memory \citep{nuxoll2007extending} storing agent's past behaviors. The switching mechanism $\lambda_t$ adaptively determines which system to rely on. For learning based switch $\lambda_t = \sigma(f_{\theta_{\lambda}}(s_t, g, h_t))$. With compromise on generalization, $\lambda_t$ can alternatively be implemented through manually designed heuristic rules. 

This mathematical insight aligns remarkably well with \citet{kahneman2011thinking}'s \textit{dual process theory of cognition}, the distinction between "thinking fast" (\texttt{System 1}) and "thinking slow" (\texttt{System 2}).   We believe this formulation reflects the fundamental cognitive principles observed in human web interaction, that is to say, \texttt{System 1} efficiently handles familiar webpage patterns (e.g., recognizing and clicking standard navigation elements), while \texttt{System 2} engages for complex scenarios such as intermittently reasoning about multi-step goals, carefully composing lengthy content replies, and handling urgent interruptions like advertisements and CAPTCHAs.

\subsection{Dual Theory Bridges Offline Imitation Learning and Online Learning}
\label{sec2.3}

Building upon our formulation of dual cognitive processes for web navigation, we examine how this theoretical framework naturally bridges offline imitation learning and online reinforcement learning paradigms. The decoupled nature provides a principled approach to leverage both learning regimes optimally. \texttt{System 1} represents the intuitive and reactive components of web navigation that can be effectively acquired through (\textbf{tactical}) offline imitation learning
\begin{equation}
\mathcal{L}_{\text{offline}}(\theta_1) = \mathbb{E}_{(s_t,g,a_t) \sim \mathcal{D}} [-\log p_{\theta_1}(a_t|s_t,g)]
\end{equation}
where $\mathcal{D}$ represents a dataset of expert trajectories. Conversely, \texttt{System 2} requires online interaction with the environment to develop adaptive (\textbf{strategic}) planning capabilities
\begin{equation}
\mathcal{L}_{\text{online}}(\theta_2) = \mathbb{E}_{(s_{1:t},a_{1:t-1},g)} [-\log p_{\theta_2}(a_t|s_{1:t},a_{1:t-1},g)]
\end{equation}
This decoupling offers advantages that \texttt{System 1} can leverage the \textit{wealth} of offline demonstrations available in benchmarks such as Mind2Web \citep{deng2023mind2web}, while \texttt{System 2} can adapt to the web's \textit{dynamic} nature through online interaction as required in WebArena \citep{zhou2023webarena}.

\textbf{Reranking Optimization.} To effectively learn \texttt{System 1}, numerous recent works have applied reranking mechanisms to improve action selection (\textit{a.k.a.} candidate generation \citep{deng2023mind2web} or dense markup ranking \citep{lu2024weblinx}). We can express $\pi_1$ proportionally as $p_{\theta_1}(a_t|s_t, g) \propto p_{\theta_1}(s_t|a_t, g) \cdot p_{\theta_1}(a_t|g)$ given that $p_{\theta_1}(s_t|g)$  is constant for all elements at a given state. This reveals two critical factors in element selection, the compatibility between the element and current state, and prior relevance of the element to the intent. We parameterize $p_{\theta_1}$ with a scoring function $f_{\theta_1}$
\begin{equation}
p_{\theta_1}(a_t|s_t, g) = \frac{\exp(f_{\theta_1}(a_t, s_t, g))}{\sum_{j=1}^{N} \exp(f_{\theta_1}(a_j, s_t, g))}
\end{equation}
For text-based representations, the scoring function $f_{\theta_1}$ can be implemented in a cross-encoder manner \citep{devlin2019bert} $f_{\theta_1}(a_t, s_t, g) = \psi_{\theta_1}([a_t; s_t; g])$, or a bi-encoder architecture $f_{\theta_1}(a_t, s_t, g) = E_{\theta_1}(a_t)^T E_{\theta_1}(s_t, g)$, where $\psi$ is a transformer-based encoder and $E_{\theta_1}$ is an embedding function.

\textbf{Visual Processing.} Current vision transformer-based approaches \citep{lee2023pix2struct, hong2024cogagent} learn representations of image patch tokens corresponding to web elements. Formally, we can express this as learning a mapping $\phi_{\theta_{1 \text{ViT}}}: \mathcal{I} \rightarrow \mathcal{R}^{d \times n}$ where $\mathcal{I}$ represents the image space, $d$ is the embedding dimension, and $n$ is the number of visual tokens. This can be integrated with the representation learning objective through
\begin{equation}
f_{\theta_1}(a_t, s_t, g) = \langle \phi_{\theta_{1 \text{ViT}}}(a_t), \psi_{\theta_{1 \text{Enc}}}(s_t, g) \rangle
\label{eq_vis}
\end{equation}
where $\psi_{\theta_{1 \text{Enc}}}$ maps the state and intent to the same embedding space. \eqref{eq_vis} establishes an equivalence between vision-based element locating and reranking HTML, providing a unified perspective on learning \texttt{System 1} capabilities across modalities.

\textbf{Preference Learning.} \citet{liu2024wepowebelementpreference} specified that the core objective for \texttt{System 1} is to learn representations that maximally discriminate target elements from other webpage elements given the current state and intent. This naturally leads to a contrastive learning objective
\begin{equation}
    \mathcal{L}_{\text{PL}}(\theta_1) = -\mathbb{E}_{(s_t, g, a_t^+, a_t^-)} \left[ \log \frac{\exp(f_{\theta_1}(a_t^+, s_t, g))}{\exp(f_{\theta_1}(a_t^+, s_t, g)) + \exp(f_{\theta_1}(a_t^-, s_t, g))} \right]
    \label{eq_wepo}
\end{equation}
which aligns with preference optimization frameworks such as WEPO \citep{liu2024wepowebelementpreference} based on \citet{rafailov2023direct}, and can be seamlessly integrated with large language models for enhanced performance. \eqref{eq_wepo} allows the model to efficiently learn to discriminate between relevant and irrelevant web UI elements (sampled from $s_t$ randomly or based on semantic distance) given the current state and intent, effectively capturing the intuitive pattern-matching capabilities of \texttt{System 1}.

\begin{figure}[htbp]
    \centering
    \includegraphics[width=0.8\textwidth]{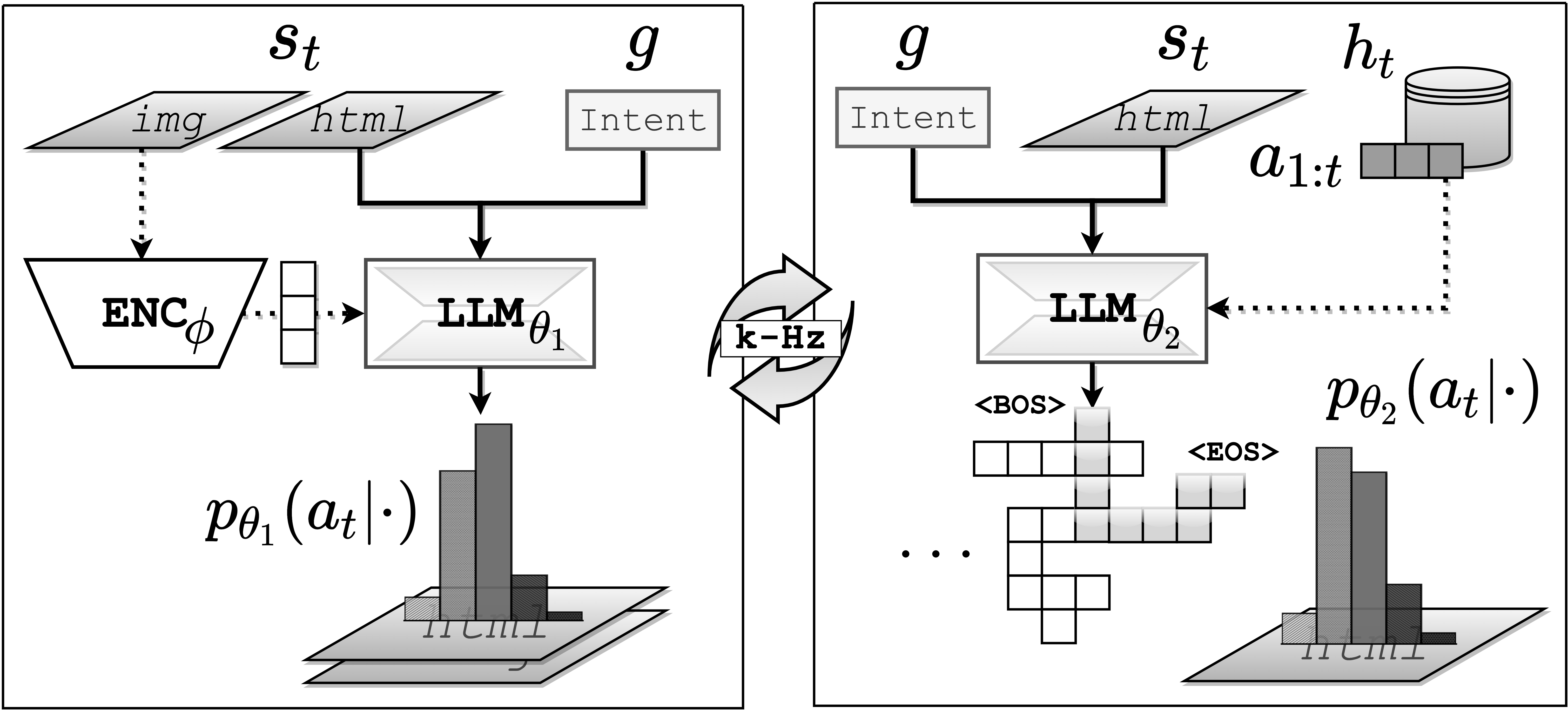}
    \caption{\texttt{CogniWeb} architecture diagram. Illustrates the web agent architecture based on dual-system cognitive theory: \texttt{System 1} on the left side and \texttt{System 2} on the right side, each receiving different combinations of state information and outputting action distributions. Dashed arrows represent optional modules, including visual alignment in \texttt{System 1} and episodic memory enhancement in \texttt{System 2}. Both systems are coordinated by the central \texttt{Switch} mechanism, which intelligently transitions between systems based on task complexity and execution status.}
    \label{fig:framework}
\end{figure}

\textbf{Online Reinforcement Learning.} \citet{qi2025webrltrainingllmweb} represents a notable attempt to address the challenges of web agent training through reinforcement learning. They formulate web navigation as a finite-horizon Markov Decision Process where the agent receives a binary reward signal at completion, and gives a KL-constrained policy update objective 
\begin{equation}
    \mathcal{L}(\theta_2) = \mathbb{E}_{(s_{1:t},a_{1:t-1},g)}\left[\left(\vartheta \log \frac{\pi_{\theta_2}(a_t|s_t, g)}{\pi_{ref}(a_t|s_t, g)} - A^*(s_t, a_t, g)\right)^2\right]
    \label{eq_webrl}
\end{equation}
where $A^*$ is the advantage and $\vartheta$ controls the divergence between current and reference policies. \eqref{eq_webrl} effectively balances retaining past knowledge while learning from new observations, addressing the distribution drift challenge inherent in curriculum learning. Similarly, \citet{prasad2024adaptasneededdecompositionplanning} employs a hierarchical planning framework where high-level policies predict subgoals for low-level policies to execute, enabling flexible decomposition of web tasks. Although not explicitly modeled within a cognitive reflection framework, these approaches lay foundation for \texttt{System 2} online learning.

\textbf{Reasoning.}\label{reasoning} For \texttt{System 2}, which handles complex scenarios requiring deliberative reasoning, we propose learning frameworks that align with recent advancements in chain-of-thought reasoning as exploration within a language model's vocabulary space \citep{yao2023tree, hao2023reasoning, shao2024deepseekmath, qin2024o1}. In fact, the theoretical derivation in Section \ref{sec2.2} naturally leads to the design of test-time reasoning. The slow nature of \texttt{System 2} corresponds to the iterative process of reasoning over multiple tokens, which requires more time but explores a much deeper space with greater potential for optimizing complex tasks. In this sense, slow thinking can be characterized by an expansive search process that traverses an intricate tree structure, leading to more thoughtful, deliberate actions that take into account long-term consequences and potential risks.

\textbf{Self-Reflection.} LLM agents are inherently \textit{stateless} \citep{sumers2023cognitive}, necessitating explicit and dynamic synchronization of knowledge updates during interaction. To address this, works like \citet{shinn2023reflexionlanguageagentsverbal} extend the policy to $\pi_2(a_t, h_{t+1} \mid s_t, h_t, g)$, where $h_t$ is no longer restricted to a simple sequence of past actions, but instead encodes rich representations of the agent's episodic experiences. This formulation allows $h_t$ to serve as a persistent medium for transmitting accumulated knowledge across otherwise stateless interactions, enabling continual adaptation and learning.

\section{Implementation}
Building upon our theoretical framework, we implement \texttt{CogniWeb}, a modular agent architecture for web navigation that operationalizes the fast-slow thinking paradigm derived in \ref{sec2.2}. As shown in Figure \ref{fig:framework}, \texttt{CogniWeb} consists of three principal components \texttt{System 1}, \texttt{System 2}, and a \texttt{Switch} mechanism that orchestrates transitions between these systems during task execution. 

\textbf{\texttt{System 1}}. We employ pretrained language models with instruction following. Our primary implementation utilizes \texttt{gpt-4o}, prompted to understand the task, interpret environmental information, and generate appropriate actions within the defined action space. We refine prompting strategy from \citet{zhou2023webarena} to elicit direct action outputs, maintaining "stop" action with unachievable (UA) hint. Although reasoning models could be integrated into \texttt{System 1}, this would contradict its inherently fast, intuitive nature. We include this configuration as an ablation.

\textbf{Offline Learning.} We also enhance \texttt{System 1} through post-training methods as described in \ref{sec2.3}. Given the theoretical equivalence of web element localization, we implement supervised fine-tuning (SFT) for reranking, using \texttt{Phi-3-mini-128k-instruct} (3.8 billions) and \texttt{gemma-3-1b-it} as base models. We construct a hybrid dataset by combining training samples from MiniWoB++ and Mind2Web, following a similar approach to \citet{lai2024autowebglm}. While the offline dataset contains websites similar to those in WebArena (e.g., e-commerce sites), the specific website contents and designs do not overlap, ensuring a rigorous evaluation of the system's generalization capabilities.

\textbf{\texttt{System 2}}. We build this module upon pretrained language models, but with specific modifications to enable complex reasoning processes. We prompt the model to produce chain-of-thought reasoning steps before generating actions, instructing it to analyze the current situation in depth with prompts \textit{"Please consider your previously executed actions and the current state, analyze the task completion progress, and identify potential errors or unnecessary historical actions..."} This reflective process enables the system to navigate complex scenarios that require strategic planning and error correction.

\textbf{Episodic Memory.} This optional component efficiently transmits information between interactions within limited context windows, helping to avoid repeating past errors. While manually engineered experiences can be incorporated through prompt engineering—a pragmatic approach in production environments—it limits generalizability. Instead, we adopt an approach inspired by \citet{shinn2023reflexionlanguageagentsverbal}, enabling the agent to generate experiential summaries based on success (or failure) metrics from current execution trajectories, creating an experience replay pool that enhances future performance.

\textbf{Working Memory.} We allow \texttt{System 2} to consider multiple previous actions ($k=10$) rather than just the most recent one. This expanded view enables the reasoning process to identify errors or repetitive patterns in longer action sequences.

\textbf{\texttt{Switch}}. While simple heuristics similar to those used in SwiftSage \citep{lin2023swiftsage} offer straightforward implementation, they face generalization challenges. We therefore implement a hybrid approach combining rule-based switching with prompt learning. We analyze performance disparities between human experts and \texttt{gpt-4o} on randomly sampled cases, constructing representative examples as few-shot demonstrations, which include \textit{switching scenarios for map search box inputs}, \textit{shopping management system information retrieval}, and \textit{special error message handling}. Theoretically, \texttt{Switch} could be further refined through SFT to achieve an optimal balance between accuracy and efficiency. We complement this learning-based approach with rule-based switching for common failure patterns. Specifically, when the agent encounters \textit{stuck} conditions (an action repeated unsuccessfully three times) or \textit{invalid} states, \texttt{CogniWeb} automatically transitions to \texttt{System 2} for deeper reflection. The pseudocode for our implementation is provided in Appendix A.

\section{Evaluation}
    \begin{table*}
    \centering
    \caption{Performance comparison of different \texttt{CogniWeb} configurations on WebArena. The results demonstrate the efficiency-accuracy trade-off between dual-system and single-system approaches, with the combined architecture achieving optimal balance.}\label{results}
    \begin{tabular}{p{0.24\textwidth}|p{0.24\textwidth}|p{0.15\textwidth}p{0.18\textwidth}}
        \toprule
        \midrule
        \texttt{System 1} & \texttt{System 2} & \textbf{Success Rate} & \textbf{Tokens per Traj.} \\
        \midrule
        \texttt{Phi-3-mini.} + SFT & \texttt{gpt-4o} + reason. + refl. & \textbf{43.96} & 393.89 \\
        \midrule
        \texttt{gemma-3-1b-it} + SFT & \texttt{gpt-4o} + reason. + refl. & 40.15 & 402.92 \\
        \midrule
        \texttt{gpt-4o} & \texttt{gpt-4o} + reason. + refl. & \underline{41.99} & 387.10 \\
        \midrule\midrule
        \textbf{Not Used} & \texttt{gpt-4o} + reason. + refl. & \textit{46.06} & 1503.83 \\
        \midrule
        \textbf{Not Used} & \texttt{gpt-4o} + reason. & 22.41 & 1421.94 \\
        \midrule
        \texttt{gpt-4o} & \textbf{Not Used} & 15.64 & 167.99 \\
        \midrule
        \bottomrule
    \end{tabular}
    \end{table*}
We conduct comprehensive experiments on WebArena's 812 tasks (2 epochs). We record both performance and efficiency metrics (task success rate and tokens per trajectory). As shown in Table \ref{results}, \texttt{CogniWeb} achieves success rates between 40\% and 45\%, outperforming mainstream approaches \citet{zhang2025webpilot,sodhi2023step} by approximately 5\% to 10\%. While our performance scores do not surpass those of \citet{zhang2025symbiotic,su2025learn}, \texttt{CogniWeb}'s streamlined algorithmic framework—which avoids excessive human engineering, experience leakage, and external knowledge bases—maintains competitive performance. Our approach's emphasis on budget-conscious efficiency and smaller but faster foundation model selection preserves better scaling potential when better data and pre-trained models become available.

\textbf{Offline learning can significantly enhance efficiency without sacrificing accuracy.} As shown in Table \ref{results}, the fine-tuned \texttt{Phi-3-mini-128k-instruct} performs best, achieving a 43.96\% success rate, slightly outperforming checkpoints based on \texttt{gemma-3-1b-it} and \texttt{gpt-4o}. In comparison, while a purely reasoning-based agent (\texttt{System 2} only) achieves a marginally higher success rate of 46.06\%, it requires nearly four times the average output length (1503.83 tokens versus 393.89) to achieve this modest 2\% performance improvement.

We observe that the current offline dataset combined with LLMs under 5B parameters provides similar benefits to \texttt{gpt-4o} with few-shot prompting. We attribute this to multiple factors: the complexity and diversity of current benchmark tasks, distribution shifts between online and offline environments, and limitations in learning efficiency. Fundamentally, WebArena's evaluation emphasizes \texttt{System 2} capabilities over \texttt{System 1}, as its websites have been distilled and simplified from real-world scenarios, lacking the complexity found in actual web pages, which demonstrate substantially higher complexity \citep{liu2024wepowebelementpreference}. 

We hypothesize that as web page complexity increases in testing environments, current algorithm performance would likely decline, making \texttt{System 1}'s pre-trained behavior cloning increasingly valuable. However, upgrading WebArena to fully represent real-world web characteristics—such as extremely long text pages and dynamic noise patterns—represents a substantial engineering challenge. The separation of \texttt{System 1} capabilities through \texttt{CogniWeb}'s framework enables offline learning prior to deployment, addressing a critical need for web navigation agents. Compared to direct behavior cloning on human trajectories, which suffers from compounding errors, our approach mitigates out-of-distribution challenges by learning from collected data and complementing this with the fast-slow thinking framework. The online learning and reasoning components keep the agent \textit{"on track"} with greater adaptability. Additionally, a unified action space definition would benefit the field, as current works employ inconsistent action space specifications \citep{drouin2024workarena}.

\textbf{Ablation Study.} We perform ablation on both \texttt{System 1} and \texttt{System 2} components, as shown in the bottom three rows of Table \ref{results}. The results demonstrate that removing either system degrades overall performance (considering both accuracy and efficiency trade-offs), validating \texttt{CogniWeb}'s dual-system design effectiveness. Additionally, we observe that the episodic memory mechanism (self-reflection) integrated with \texttt{System 2} provides substantial performance benefits, highlighting the advantages of our modular system design. When this elegant inter-interaction information transmission component is removed from \texttt{System 2}, performance precipitously drops from approximately 40\% to 20\% success rate.

\begin{figure}[htbp]
    \centering
    \includegraphics[width=0.8\textwidth]{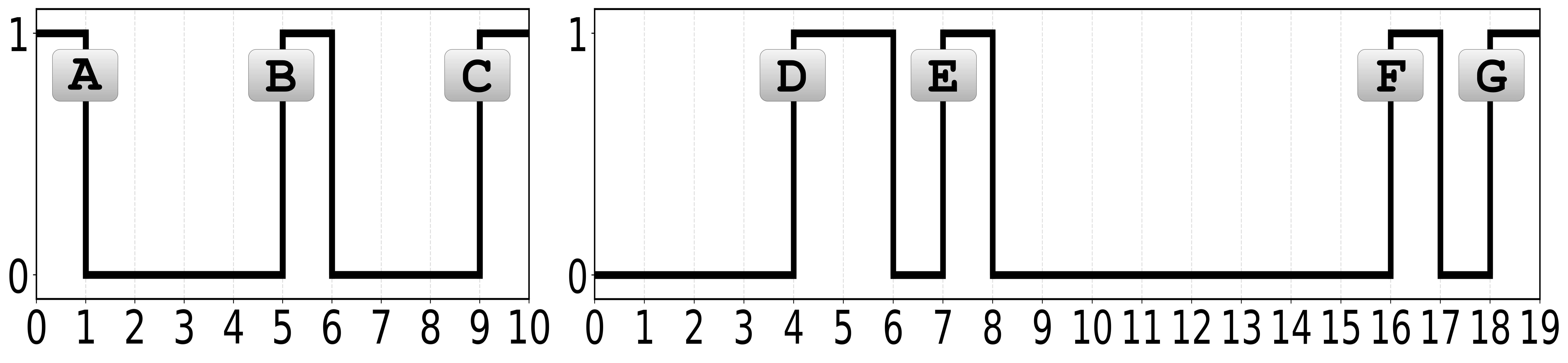}
    \caption{Qualitative analysis of \texttt{CogniWeb}'s dual systems. The figure illustrates two examples where \texttt{CogniWeb}'s dual systems successfully collaborate to solve complex tasks (task 368 and task 27). The figure resembles level switching in electronic circuits, with the y-axis value 0 corresponding to \texttt{System 1} and value 1 corresponding to \texttt{System 2}.}
    \label{fig:switching}
\end{figure}

\textbf{Qualitative Analysis.} Figure \ref{fig:switching} illustrates two examples where \texttt{CogniWeb}'s dual systems successfully collaborate to solve complex tasks. Task 368, with the intent \textit{"find discounted items,"} demonstrates the complementary strengths of both systems. At point A, \texttt{System 2} initially takes control for planning, followed by \texttt{System 1} executing exploratory clicks from A$\rightarrow$B. At point B, \texttt{System 2} reactivates, reasoning through verification strategies: \textit{"checking whether there is a discount filter option"} and \textit{"I could use the search box to search for the keyword discount, but this seems impractical as the results may not have temporal relevance."} This strategic thinking successfully navigates a key obstacle, after which \texttt{System 1} completes the task through a series of scrolling and clicking actions. Points C and G represent \texttt{System 2}'s experience summarization for the replay pool.
Task 27 presents an even more complex challenge: \textit{"Tell me the count of comments that have received more downvotes than upvotes for the user who made the latest post on the Showerthoughts forum."} At point D, \texttt{System 2} initiates a new tab and attempts to utilize the wiki tool. Finding this unproductive, \texttt{System 2} reflects at point E and decides to return to the original page, leveraging the forum interface rather than persisting with the new tab. From E$\rightarrow$F, \texttt{System 1} rapidly handles navigation interactions until point F, where the system thoughtfully considers: \textit{"I should double-check whether there are filtering functions like find more to ensure I haven't missed any comments,"} ultimately confirming the answer as 0.

\section{Related Work}
Given our comprehensive review of web-based agents in Section \ref{sec:theory}, we focus here on dual-system cognitive theories in artificial intelligence. \citet{kahneman2011thinking} dual-process theory distinguishes between fast, intuitive "System 1" processes and slow, deliberate "System 2" processes—a framework increasingly applied across AI domains. Recent implementations include DynaThink \citep{pan2024dynathinkfastslowdynamic}, which enables LLMs to dynamically select between fast and slow inference methods; Fast-Slow-Thinking \citep{sun2025fastslowthinkingcomplextasksolving}, which applies coarse-to-fine problem solving; and \citet{Qi_2024_CVPR}'s Interactive Continual Learning framework that pairs smaller vision models (System 1) with larger LLMs (System 2). In cognitive architectures, Soar \citep{laird2019soar} and CoALA \citep{sumers2023cognitive} implement human-like memory systems and decision procedures, while SwiftSage \citep{lin2023swiftsage} employs dual-system theory specifically for interactive reasoning tasks. Our work represents the first principled application of dual-process theory to web navigation, uniquely bridging offline imitation learning and online exploration through a mathematical formalization of web navigation complexity.

\section{Conclusion}
In this paper, we introduced a theoretical framework that formalizes web navigation through the lens of dual-process cognitive theory, deriving a principled decomposition into fast intuitive processing (\texttt{System 1}) and slow deliberative reasoning (\texttt{System 2}). Our implementation, \texttt{CogniWeb}, demonstrates that this approach effectively balances performance and efficiency, achieving competitive success rates on WebArena. By bridging the gap between offline imitation learning and online exploration, our framework unifies diverse methodologies in current web agent research and establishes clearer pathways for future development. As web environments continue to evolve in complexity, this dual-system offers a scalable foundation for building increasingly capable autonomous agents that can efficiently navigate the web while maintaining robust performance across diverse tasks.

\bibliographystyle{plainnat}
\bibliography{references}

\newpage
\appendix
\section{Pseudocode}  
\begin{algorithm}
    \caption{CogniWeb}
    \label{alg:cogniweb}
    \begin{algorithmic}[1]  
    
    \State \pycomment{Initialize the agent and environment}
    \State \texttt{args = config(agent\_name="CogniWeb")} 
    \State \texttt{prepare(args)}
    \State \texttt{system\_1, system\_2, switch = construct\_agent(args)}
    \State \texttt{system\_2.episodic = []} \pycomment{Optional, for episodic memory}
    \State
    \State \texttt{while True:} \pycomment{Main loop in one episode}
    \State \hspace{1em}\texttt{early\_stop\_flag, stop\_info = early\_stop(trajectory, max\_steps, ...)}
    \State \hspace{1em}\texttt{if early\_stop\_flag:}
    \State \hspace{2em}\texttt{if "max steps" not in stop\_info:}
    \State \hspace{3em}\texttt{action = system\_2.next\_action(trajectory, intent, meta\_data)}
    \State \hspace{2em}\texttt{else:}
    \State \hspace{3em}\texttt{action = create\_stop\_action(f"Early stop:\{stop\_info\}")}
    \State \hspace{1em}\texttt{else:}
    \State \hspace{2em}\texttt{try:}
    \State \hspace{3em}\texttt{index = switch.switch(trajectory, intent, meta\_data)}
    \State \hspace{3em}\texttt{if index == 0:} \pycomment{thinking fast}
    \State \hspace{4em}\texttt{action = system\_1.next\_action(trajectory, intent, meta\_data)}
    \State \hspace{3em}\texttt{elif index == 1:} \pycomment{thinking slow}
    \State \hspace{4em}\texttt{action = system\_2.next\_action(trajectory, intent, meta\_data)}
    \State \hspace{2em}\texttt{except ValueError as e:} \pycomment{get the error message}
    \State \hspace{3em}\texttt{action = create\_stop\_action(f"ERROR: \{str(e)\}")}
    \State
    \State \hspace{1em}\texttt{trajectory.append(action)}
    \State \hspace{1em}\texttt{action\_str = get\_action\_description(action, ...)}
    
    \State \hspace{1em}\texttt{meta\_data["action\_history"].append(action\_str)}
    
    \State \hspace{1em}\texttt{if action["action\_type"] == ActionTypes.STOP:}
    \State \hspace{2em}\texttt{break}
    \State
    \State \hspace{1em}\texttt{obs, \_, terminated, \_, info = env.step(action)}
    \State \hspace{1em}\texttt{state\_info = \{"observation":obs, "info":info\}}
    \State \hspace{1em}\texttt{trajectory.append(state\_info)}
    \State \hspace{1em}\texttt{if terminated:} \pycomment{add a action place holder}
    \State \hspace{2em}\texttt{trajectory.append(create\_stop\_action(""))}
    \State \hspace{2em}\texttt{break}
    \State
    \State \hspace{1em}\texttt{evaluator = evaluator\_router(config\_file)}
    \State \hspace{1em}\texttt{score = evaluator(trajectory, ...)}
    \State \hspace{1em}\texttt{scores.append(score)}
    \State
    \State \hspace{1em} \pycomment{summarize the experience based on the result}
    \State \hspace{1em}\texttt{experience = system\_2.summarize(trajectory, intent, meta\_data, score)}
    \State \hspace{1em}\texttt{system\_2.episodic.append(experience)}

    \end{algorithmic}
    \end{algorithm}



\end{document}